# Transfer Learning for Risk Classification of Social Media Posts: Model Evaluation Study

Authors: Derek Howard[1,2], Marta Maslej[1], Justin Lee[3], Jacob Ritchie[1,4], Geoffrey Woollard[5,6], Leon French[1,2,7,8]

Affiliations
[1] Campbell Family Mental Health Research Institute, Centre for Addiction and Mental Health, Toronto, Canada
[2] Krembil Centre for Neuroinformatics, Centre for Addiction and Mental Health, Toronto, Canada
[3] Department of Biochemistry, University of Toronto, Toronto, Canada
[4] Department of Computer Science, University of Toronto, Toronto, Canada
[5] Department of Medical Biophysics, University of Toronto, Toronto, Canada
[6] Princess Margaret Cancer Centre, University Health Network, Toronto, Canada
[7] Institute for Medical Science, University of Toronto, Toronto, Canada
[8] Division of Brain and Therapeutics, Department of Psychiatry, University of Toronto, Canada
Corresponding author: Leon French (leon.french@camh.ca)

## Abstract

**Background**
Mental illness affects a significant portion of the worldwide population. Online mental health forums can provide a supportive environment for those afflicted and also generate a large amount of data which can be mined to predict mental health states using machine learning methods.

**Objective**
We benchmark multiple methods of text feature representation for social media posts and compare their downstream use with automated machine learning (AutoML) tools to triage content for moderator attention.

**Methods**
We used 1588 labeled posts from the CLPsych 2017 shared task collected from the *Reachout.com* forum (Milne et al., 2019). Posts were represented using lexicon based tools including VADER, Empath, LIWC and also used pre-trained artificial neural network models including DeepMoji, Universal Sentence Encoder, and GPT-1. We used TPOT and auto-sklearn as AutoML tools to generate classifiers to triage the posts.

**Results**
The top-performing system used features derived from the GPT-1 model, which was finetuned on over 150,000 unlabeled posts from *Reachout.com*. Our top system had a macro averaged F1 score of 0.572, providing a new state-of-the-art result on the CLPsych 2017 task. This was achieved without additional information from meta-data or preceding posts. Error analyses revealed that this top system often misses expressions of hopelessness. We additionally present visualizations that aid understanding of the learned classifiers.

**Conclusions**
We show that transfer learning is an effective strategy for predicting risk with relatively little labeled data. We note that finetuning of pretrained language models provides further gains when large amounts of unlabeled text is available.

## Introduction

Depression is the worldwide leading cause of disability, and a similar number of people suffer from a range of anxiety disorders. The number of people with mental health disorders is increasing globally, and anyone can be affected at any time in their life [1].

Given the high incidence of mental health disorders and the relatively low incidence of self-harm, predicting risk for self-harm is very difficult. In particular, Franklin et al. report a lack of progress over the last 50 years on the identification of risk factors that can aid in the prediction of suicidal thoughts and behaviours [2]. However, they also propose that new methods with a focus on *risk algorithms* using machine learning present an ideal path forward. These approaches can be integrated into peer-support forums to develop repeated and continuous measurements of a user's well-being to inform early interventions.

Peer support forums can be a useful and scalable method of social therapy for mental health issues. Many individuals are already seeking health information online, and this method of information access can help those that are reluctant to seek professional help, are concerned about stigma or confidentiality, or face barriers to access [3]. There is limited evidence showing that online peer-support without professional moderation is an effective strategy for enhancing users' well-being [4]. However, in a systematic review of social networking sites for mental health interventions, [5] identify the use of moderators as a key component of successful interventions on these online platforms [5]. The development of automated triage systems in these contexts can facilitate professional intervention by prioritizing users for specialized care [6,7] or improving the rate of response when a risk for self-harm is identified [8].

Previous research suggests that the language of individuals with mental health conditions is characterized by distinct features [9–12]. For example, frequent use of first person singular pronouns has been associated with depression [13]. This has enabled the development of systems that can classify the level of user risk from passively collected social media data. Extracting features that best represent the social media text is a key step of such systems. A broad array of language representation and classification methods have been applied to triage on posts from the Australian mental health forum *Reachout.com*, which provides mental health information and support for youth online [14,15]. Recently, transfer learning has been used for text classification across a variety of contexts. The fine-tuning of pretrained word embeddings has flourished in applied settings and can take advantage of unlabeled data [16]. More recently, pretrained word representations have shown the ability to better capture complex contextual word characteristics [17] and the fine-tuning of large pre-trained language models in an unsupervised fashion has pushed forward the applicability of these methods in cases with small amounts of labeled data [18,19]. In the present paper, we benchmark feature extraction methods on forum posts from *Reachout.com*. We show that modern transfer learning approaches, in combination with automated machine learning tools, can train highly performing triage systems.

# Methods

## Data

### Reachout.com

Our primary data source was made available for the CLPsych Shared task 2017 and was collected from the Australian mental health peer-support forum *Reachout.com [8,14]*. The entire dataset consists of 157,963 posts written between July 2012 and March 2017. Of those, 1188 were labeled and used for training the classification system, and 400 labeled posts were held out for final evaluation of the systems. Posts were labeled 'green' (58.6%), 'amber' (25.6%), 'red' (11.7%) or 'crisis' (5.2%) based on the level of urgency with which moderators should respond. The annotators followed a flowchart/rubric/decision tree to standardize the labeling process and included fine-grained or granular annotations for each of the posts (Supplementary Table S1).

### UMD Reddit Suicidality Dataset

To test the external validity of the system developed on the Reachout.com data, we used a subset of the data made available from the *UMD Reddit Suicidality Dataset [20,21]*. Collection of this dataset followed an approach where the initial signal for a positive status of suicidality was a user having posted in the subreddit */r/SuicideWatch* between 2006-2015. Annotations were then applied at the user level based on their history of posts. The annotations fall into four categories ('a': No risk, 'b': Low risk , 'c': Moderate risk, 'd': High risk) which roughly correspond to the *green, amber, red, crisis* categories defined for the *Reachout.com* data. We used the subset that was curated by *expert* annotators to assess suicide risk. Of the subset with labels by expert annotators, we then selected only data from users who had posted a single time in */r/SuicideWatch* to minimise ambiguity in understanding which of their posts was the cause of the associated label. Predictions were made only on posts from */r/SuicideWatch*. In total, there were 179 user posts across the categories ('a': 32, 'b': 36, 'c': 85, 'd':26). The Centre for Addiction and Mental Health Research Ethics Board approved the use of this dataset for this study.

To better benchmark our performance on the UMD Reddit Suicidality Dataset posts, we calculated a random baseline for the macroF1 metric. This baseline evaluated the performance of random shuffles of the true labels(including the class 'a' or 'no risk' labels). Across 10,000 of these randomizations, the mean macroF1 is 0.25. Corresponding to a *P* > 0.05 and Bonferonni correction for eight tests, we set a threshold of 0.336 which is 62 of 10,000 random runs to mark Reddit validation performance as better than chance (1/20 × ⅛ × 10000).

### Composite quotes

We used ten composite quotes to share example predictions of our system on text that could be predictive/indicative of self-harming and/or suicidality. These composite quotes were created by Furqan et al. [22], and were derived from qualitative research that synthesized primary themes noted in a selection of suicide notes that made explicit mentions of mental illness and/or mental healthcare. To assess the role of individual words (or tokens) in the classification of the quote, we iteratively perturbed each token and replaced it with an unknown token outside of the model's vocabulary and re-ran the prediction.

### Features

#### Data pre-processing and feature extraction

Features were extracted from only the text body of the posts. For all posts, any quote of a previous post ,was removed from the post which contained it, and links to images were removed.

We extracted features using lexicon based tools such as VADER (4 features) [23], LIWC (70 features) [24] and Empath (195 features) [25] which have proven to be useful for characterizing social media text and extracting psychologically relevant signals. Features were also extracted from three pre-trained artificial neural network models: DeepMoji [26] was used to extract sentiment and emotion-related features (e.g., the use of emoticons in social media text); the Universal Sentence Encoder [27] v2 (using a Deep Averaging Network encoder) obtained from Tensorflow Hub which was specifically designed to facilitate transfer learning; and the Generative Pre-Trained network (GPT) v1 developed by OpenAI [19]. For DeepMoji, we extracted features that represent the 64 predicted emoji and also the neural activations from the preceding attention layer in the network (2,304 features, referred to as DeepMoji). We used the Indico Data Solutions implementation available at https://github.com/IndicoDataSolutions/finetune to extract features from the default pre-trained GPT-1 network and also after finetuning on the unlabeled corpus of posts from Reachout.com. All language model finetuning was done with three epochs over the unlabeled posts as suggested by the GPT-1 authors.

With Empath and LIWC, sentence splitting was not performed. With the remaining feature encoding (VADER, Deepmoji, Universal sentence encoder and both GPT models) methods, we first pre-processed the text body of each post into sentences using the sentence boundary detection from spaCy v2.1 [28]. Sentence feature vectors were aggregated to the post level by taking their mean, max and min for each extracted feature.

## Model optimization and selection

To train classifiers on the various feature sets, we used two automated machine learning methods (AutoML) that are built upon scikit-learn [29] to optimize and select optimal models. In both cases, the AutoML methods were customized to maximize the Macro-F1 score (without the 'green' labelled posts). The organizers of the CLPsych shared task chose this metric to weight each class for both precision and recall equally. Each model was evaluated with 10-fold stratified cross-validation with five repeats inside of the training set. We trained the classifiers to predict the granular/fine-grained labels while evaluating the final output with the same Macro-F1 score of the 'amber', 'red' and 'crisis' categories.

We used TPOT, the Tree-based Optimization Tool [30], which builds and selects machine learning pipelines using genetic programming. TPOT is built to generate pipelines which maximize classification accuracy while penalizing complex pipelines. Similarly, we used AutoSKlearn to train and build classifiers using Bayesian optimization meta-learning and ensemble construction [31]. We primarily used default TPOT/auto-SKlearn parameters with a population size of 200, a max evaluation time for a single pipeline of 5 minutes and total time as a stopping parameter, typically set to 2 days.

### Mantel tests

To compute the matrix of pairwise Euclidean distances between posts for each set of features we used SciPy's distance_matrix function [32]. This test allows quantification of the distances between posts across the various feature spaces. This is done in an unsupervised manner across the training and test posts. We used scikit-bio's mantel function with 999 permutations to perform the Mantel test on these distance matrices (999 permutations).

### Emoji visualization

To better understand the distribution of the 64 emoji features represented across the labelled posts, we aggregated the mean of an emoji feature across sentences in a post. Each of these aggregate features was then normalized to the (0,1) range to better compare features against each other. To obtain a measure of feature importance, we permuted each feature column and assessed the decrease in classification performance on the macroF1 metric while using the best performing pipeline derived from TPOT. For each emoji feature, we performed this procedure 10,000 times. Images of the emoji's were obtained from emojione.com and converted to grayscale (currently joypixels.com).

### Availability

The CLPsych 2017 and UMD Reddit Suicidality datasets are available upon request from the original sources [20,21]. Code and instructions to finetune, train and test a GPT-1 model on the CLPsych 2017 dataset is available online at: https://github.com/derekhoward/Reachout_triage

## Results

To benchmark the performance of various text derived features for the automated classification of online forum posts, we ran both TPOT and AutoSklearn on the features generated from the posts bodies. In Table 1, we report the average observed score across the training folds, the final score on the held out Reachout.com test set, and the external validation performance on Reddit data of the classifier trained only on Reachout.com data.

We note that the average macroF1 obtained during training is a fairly reliable predictor of the score on the held out test set. AutoSklearn performed better on average than TPOT (mean test MacroF1 of 0.414 versus 0.379). We also observe the trend that features extracted from pretrained models perform better in general (average AutoSklearn MacroF1 of 0.329 versus 0.466). However, the features extracted from the default GPT model (without any additional fine-tuning) were the worst performing of those obtained from neural models, while the GPT model that was finetuned on the unlabelled posts performed best across all experiments. The Universal Sentence Encoder and finetuned GPT features exceeded the highest macroF1 score reached in the CLPsych 2017 shared task when a classifier is learned with AutoSklearn (0.467, Xianyi Xia and Dexi Liu). Upon inspection, the AutoSklearn generated classifier for the GPT finetuned features are complex ensembles of pipelines with multiple preprocessing steps and random forest classifiers. The TPOT generated classification pipeline first selects features using the ANOVA F-value, then binarizes the values for classification with a K-nearest neighbour classifier (k=21, Euclidean distance). In contrast, the classifiers generated for the Universal Sentence Encoder features are a linear support vector machine (TPOT) and ensembles of linear discriminant analysis classifiers (AutoSklearn).

To better understand the low Reddit validation scores, we calculated a random baseline. While random, this does use information about the class distributions. We mark Reddit validation performance as better than chance in Table 1 with a superscript ([a]). Only classifiers learned from the Vader, DeepMoji, and default GPT features had macroF1 scores above the threshold for both the TPOT and AutoSklearn learned classifiers. Unlike the CLPsych 2017 score that does not include the 'green' or no risk labels, we used macroF1 from all classes in the Reddit validation tests (corresponding to the CLPsych 2019 primary metric). When using the macroF1 score that excluded the 'no risk' class in the Reddit validation, none of the classifiers outperformed random runs at the same threshold. This is due to the classifiers having good performance on the 'no risk' or 'green' labels and not the three remaining labels.

**Table 1. Benchmarking by features, AutoML method, and datasets with the macroF1 metric.**

| | | TPOT | | | AutoSklearn | | |
|---|---|---|---|---|---|---|---|
| | Feature count | train 10-fold 5x | test | Reddit validation | train 10-fold 5x | test | Reddit validation |
| **Empath (post)** | 195 | 0.280 | 0.253 | 0.385 [a] | 0.292 | 0.344 | 0.321 |
| **LIWC** | 70 | 0.434 | 0.354 | 0.346 [a] | 0.433 | 0.380 | 0.315 |
| **Vader (sentence)** | 12 | 0.363 | 0.263 | 0.356 [a] | 0.340 | 0.263 | 0.353 [a] |
| **Emoji 64** | 192 | 0.425 | 0.369 | 0.280 | 0.424 | 0.461 | 0.308 |
| **DeepMoji** | 6912 | 0.442 | 0.452 | 0.345 [a] | 0.391 | 0.437 | 0.351 [a] |
| **Universal Sentence Encoder** | 1536 | 0.457 | 0.446 | 0.300 | 0.484 | 0.479 | 0.236 |
| **GPT-default** | 2304 | 0.373 | 0.334 | 0.344 [a] | 0.396 | 0.383 | 0.402 [a] |
| **GPT-finetuned** | 2304 | 0.510 | 0.559 | 0.320 | 0.492 | 0.572 | 0.324 |

To better assess the variability of our best performing system (AutoSklearn trained with features generated from the fine-tuned GPT model), we re-ran the AutoSklearn training and testing process 20 times. For each run, AutoSklearn was allotted 24 hours of compute time. Across those 20 systems, the average macroF1 score on the held out test set was 0.5293 with a standard deviation of 0.0348. Of those 20, the best and worst performing systems had a final test score of 0.6156 and 0.4594, respectively. Importantly, despite the variability and less compute time, the average macroF1 score of these classifiers performed better than the scores obtained from different feature sets.

To determine the impact of the amount of data used for fine-tuning the GPT model on its effectiveness for feature extraction in the classification task, we finetuned models with increasing amounts of unlabelled posts before extracting post-level features to train a classifier (Figure 1). Although there is significant variability, there is a general trend of better performance when using models trained on a larger amount of unlabelled data.

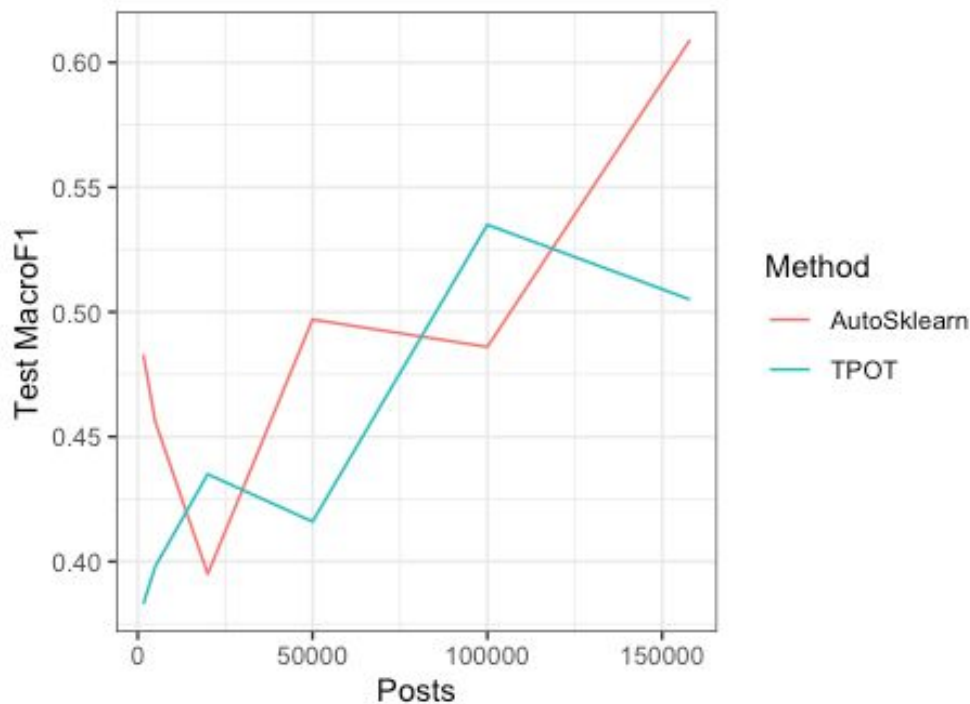

Figure 1. Graph of macroF1 test scores versus number of posts used for GPT-1 finetuning. AutoSklearn methods are marked with red (AutoSklearn) and blue (TPOT) lines.

To compare the different representations or embeddings of the post contents, we used the Mantel test. This test correlates the pairwise distances between posts in the benchmarked feature spaces. Intriguingly, we observe the highest correlation between the Universal Sentence encoded features with those encoded by GPT. This is despite the comparison of aggregated DeepMoji encoded features with aggregated 64-dimensional emoji encoding of DeepMoji which we expected to have the strongest relationship. Similarly, comparisons between the default GPT and the fine-tuned version are slightly lower than correlations with the Universal Sentence Encoder. While unclear, we presume some of these differences may be due to the aggregation of sentence-level features into a post-level representation. None of the correlations with Empath features were significant, which probably reflects the sparsity of these features.

**Table 2. Mantel correlations between the extracted feature sets.**

|  | VADER | Empath | LIWC | Universal Sentence | Emoji (64) | DeepMoji | GPT default | GPT finetuned |
|---|---|---|---|---|---|---|---|---|
| **VADER** |  |  |  |  |  |  |  |  |
| **Empath** | 0.003 |  |  |  |  |  |  |  |
| **LIWC** | 0.098 | 0.009 |  |  |  |  |  |  |
| **Universal Sentence** | 0.453 | 0.006 | 0.148 |  |  |  |  |  |
| **Emoji 64** | 0.211 | -0.005 | 0.403 | 0.193 |  |  |  |  |
| **DeepMoji** | 0.422 | -0.008 | 0.507 | 0.509 | 0.523 |  |  |  |
| **GPT default** | 0.430 | 0.004 | 0.267 | 0.823 | 0.302 | 0.632 |  |  |
| **GPT finetuned** | 0.429 | 0.001 | 0.253 | 0.823 | 0.335 | 0.631 | 0.799 |  |

## System interpretability

In Figure 2, we show the distribution of the mean emoji features for the top 10 most important features when using the mean emoji feature across sentences (64 total features). We note that the interpretation and even visual representation of these emoji varies greatly, and these emoji were not used in the social media posts but are extracted by DeepMoji [26]. For example, the pistol emoji has been replaced by a raygun or water gun in most platforms. From these distributions, it is clear that there is considerable variability across posts. This visualization also highlights the difficulty in discriminating the varying levels of risk when compared to the no risk posts. Of these top ten, two winking emoji are negatively correlated with risk, marking the importance of positive sentiment. As expected, the negative emoji are more important with pistol, skull and broken heart emoji ranked in the top five.

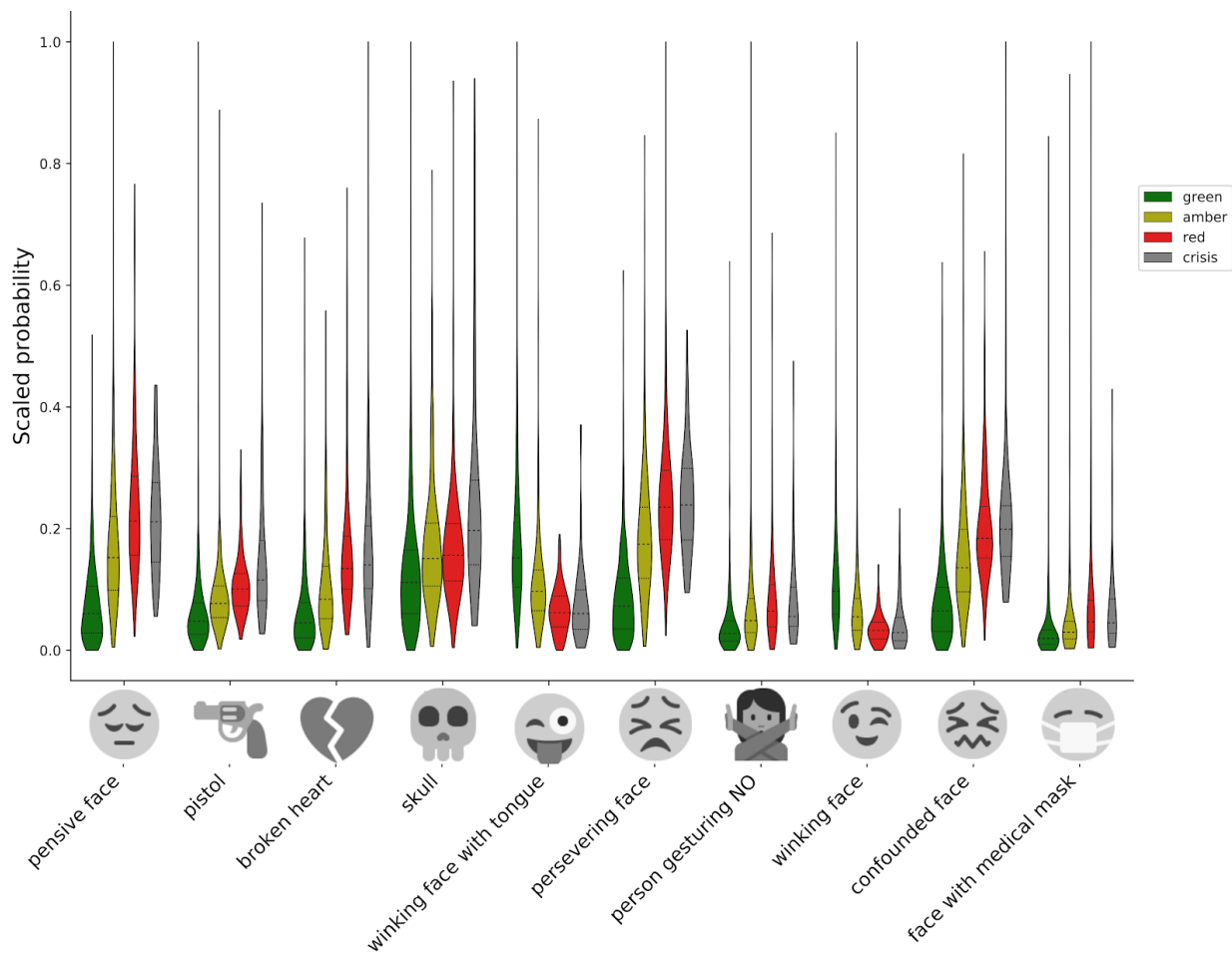

Figure 2. Violin plot showing the distributions of the ten most discriminative emoji features across labelled classes. The classes are according to label with crisis in grey. The y-axis is the predicted scores for each emoji and have been scaled to the 0-1 interval. The emoji across y axis are marked with their images and their official unicode text labels. The emoji are ranked from the most to least important feature (left-right).

## Error analysis

We manually reviewed the errors made by the best-performing system (AutoSklearn classifier with the GPT fine-tuned features). The most worrisome prediction errors occur when the classifier mistakes a crisis post for one of lesser importance, which could potentially delay a moderator response. When posts were not classified as crisis posts (but should have been), this was often due to vague language referring to self-harm or suicide (e.g., "time's up", "get something/do it", "to end it", "making the pain worse"). Sometimes forum users deliberately refer to self-harm or suicide with non-standard variations, such as "SH" or "X" (eg: "attempt X", "do X"). Future work could be instructive in determining whether these words are associated with greater levels of distress/crisis relative to the words they are meant to replace. Alternatively, custom lexicons might be developed to capture instances of self-harm or suicide represented by vague language or non-standard variations.

In some failure cases (i.e., posts that should be classified as being of higher risk than they were), the classifier did not pick up on expressions of hopelessness, which may cue the imminence of risk. Other prominent failure cases were instances when the classifier did not pick up on a poster's dissatisfaction with mental health services that provide real-time help (eg: Suicide call-back services, crisis helplines, etc). According to the labeling scheme, these posts should be classified as red. However, this dissatisfaction was often conveyed in diverse and highly contextualized ways, likely making it difficult for the system to identify. There were also posts that did not indicate imminent risk, but described sensitive topics such as feeling lonely, or losing a parent. These were often misclassified as green (when they should have been amber), possibly because they also contained positive language or the sensitivity of the topic was difficult for the system to grasp.

In some of these failure cases, it may have been useful to take into account the previous post. For example, when the post in question is short or vague, the system may classify the level of risk more accurately if the previous post expresses a high level of concern about the poster, or tries to convince the poster to seek immediate help.

To better understand judgments made by our trained classifier, we share predictions in Table 3 on a set of composite quotes and their themes from a study of suicide notes [22]. For each quote, we share the initial prediction (with the granular/fine-grained prediction in parentheses). Across the ten quotes, three are classified as crisis, four as red, and three as amber. One of the three amber classifications is under the "Hopelessness secondary to chronicity of illness and treatment" theme, further suggesting that our system may not recognize expressions of hopelessness.

All words were iteratively masked to indicate their effects on the predicted class (see Methods). In Table 3, words that affected predictions are colour-coded. Replacing a yellow or red word with an unknown word shifts the prediction to a less severe class by one or two levels respectively (i.e., replacing a yellow word shifts the predicted class from 'crisis' to 'red', replacing a red word shifts the predicted class from 'crisis' to 'amber'). These words are important for indicating severity, since removing them makes the quotes appear less severe to our system. Examining these words suggests that negations affected severity (e.g., "not", "can't"). In the quotes, negations seemed to indicate a perceived failure, or not having done or achieved something the person felt they ought to. Contrary to observations from our error analysis, expressions of hopelessness (i.e., "no hope left") were also important in classifying quotes as severe by our system. Words reflecting an unwillingness or inability to continue were also important (i.e., "I'm done", "I am too tired to"), as well as words indicating loneliness (i.e., "being isolated").

In contrast, replacing a green word with an unknown word shifted the predicted class to a more severe category (e.g., from 'red' to 'crisis'). Examining the nature of the green words (i.e., "what", "after"), it's not clear why these words were important for lessening the severity of the quotes.

For two of the quotes predicted as red, no words were highlighted, suggesting that in these instances, many words were key to the prediction. Overall, the quotes would all be flagged as requiring some level of moderator attention, and for the most part, the nature of words that were important in classifying the severity of quotes made conceptual sense.

**Table 3. Predictions and highlights of suicide related composite quotes from Furqan et al.**

| Theme | Prediction | Composite quote |
|---|---|---|
| Suicide as Exertion of Personal Autonomy | Red (currentAcuteDistress) | I am the person who decides what I m going to do with my life. After being isolated and marginalized, I m done with it all. This is my life and my decision. |
| Exhaustion | Red (currentAcuteDistress) | I am exhausted from trying to fix everything. No one understands, not even doctors or my family, and I keep trying to get help. I feel like I m a dead man walking for a long time. I ve been judged by society and have been made to feel ashamed. I m burned out and tired and trying to find some way to rest. |
| Treatment failure as personal failure | Crisis | It does n t matter how many treatments I try , there is something so wrong with me. It s not the treatments, it s me. |
| Treatment failure as personal failure | Crisis | Medication, therapy, counseling, alcohol I ve tried everything and I ca n t seem to figure out a way out of this, I ca n t crack the code. |
| Hopelessness secondary to chronicity of illness and treatment | Amber (followupOk) | After trying multiple treatments , without curing my mental illness, I have realized that a solution other than death just does not exist. |
| Conflict between Self and Illness | Crisis | I have fought against my thoughts, depression and alcohol constantly. I am too tired to keep going |

| Illness as biological | Amber (followupOk) | There is a problem in my brain. I think the chemistry is all wrong. No one could have fixed that. |
| --- | --- | --- |
| Low levels of perceived agency | Amber (followupOk) | Forgive me. It s not my fault, it s a disease, I hope you can understand that. |
| Low levels of perceived agency | Red (followupWorse) | I sank into a deep depression. It was the depression that ruined my relationship. I want my family to know that. |
| High levels of perceived agency | Red (currentAcuteDistress) | I had so much in my life family, friends , career but I let the disease, addiction, and my own personality take over me and it ruined everything in my life. Looking back, there were times that I should have changed the course of my life but I did n t and now there is no hope left. |

## Discussion

We show that there are highly informative signals in the text body alone of posts from the *Reachout.com* forum. More specifically, we identify a transfer learning approach as particularly useful as the feature extraction phase from raw social media text. In combination with the training of classifiers using AutoML methods, we show that these representations of the post content can improve triage performance without taking into account the context or metadata of the posts. These methods take advantage of the large amount of unlabeled free-text that are often available to diminish the need for labelled examples. We also show that these methods can generalize to new users on a support-forum, for which there would not be preceding posts to provide context on their mental states. By combining the pretrained language models with AutoML, we were able to achieve state-of-the-art macroF1 on the CLPsych 2017 shared task. Our content-only approach could be complemented by previous work which used hand-engineered features to account contextual information such as a user's post history or the thread context of posts [15,33]. Future developments could also include multiple types of media (text, photos, video) that are often present on social media to better assess the subtleties of users' interactions [34].

Our current approach follows methods outlined by [19] to finetune the language model that was previously pre-trained on a large corpus of books. This finetuning allows step learns the characteristics of the text on *Reachout.com*. We show that increasing amounts of in-domain unlabeled data for finetuning improves classification performance and has yet to reach a plateau. Further work will be instrumental in defining when and how to finetune pretrained language models [35]. For tasks with limited data availability, the ability to adapt and finetune a model on multiple intermediate tasks could be a particularly worthwhile approach as demonstrated by the Universal Sentence Encoder and others [27,36]. However, it's unclear how these large language models can retain and accumulate knowledge across tasks and datasets. Notably, it has been reported that these large pretrained language models are difficult to finetune and that many random restarts can be required to achieve optimal performance [37,38].

We compared the use of AutoML tools such as AutoSklearn and TPOT to generate classification pipelines with a variety of features extracted from free text. We also identified them as sources of variability in the final scores of our system. When developing our top performing systems with features extracted from a finetuned GPT and using auto-sklearn, on 20 trials, we obtained macro average F1 scores ranging from 0.6156 to 0.4594. In part, this is due to the small size of the dataset and the weighted focus of the macro average F1 metric towards the 'crisis' class with relatively fewer instances. Further experiments, while computationally intensive, could help distinguish the amount of variability that is inherent in the language model finetuning process.

Our classifiers generalize to some degree on the UMD Reddit Suicidality Dataset which approximates the task outlined for *Reachout.com*. This performance is primarily driven by performance on the 'no risk' or 'green' class. We observe that the features derived from the finetuned GPT model perform worse than those from the default GPT model indicating that this model might be overfitting unique features of the Australian posts from *Reachout.com*. Future studies could determine whether multiple rounds of finetuning on different datasets increase accuracy.

Neural networks can build complex representations of their input features, and it can be difficult to interpret how these representations are used in the classification process. In a deeper analysis of DeepMoji features, we identified the most important emoji for classification and found that the emotional features follow a linear arrangement of expression at the class level corresponding to label severity. We also used input masking to iteratively highlight the contributions of individual words to the final classification. Such highlighting and pictorial/emoji visualizations could speed moderator review of posts. Ultimately, we believe the further development of methods to improve model interpretability will be essential in facilitating the work of mental health professionals in online contexts.

In conclusion, we show that transfer learning combined with AutoML provides state-of-the-art performance on the CLPsych 2017 triage task. Specifically, we found that an AutoML classifier trained on features from a finetuned GPT language model was the most accurate. We suggest this automated transfer learning approach as the first step to those building natural language processing systems for mental health due to the ease of implementation. While such systems lack interpretability, we show that emoji based visualizations and masking can aid explainability.

## Acknowledgements

The CAMH Specialized Computing Cluster, which is funded by The Canada Foundation for Innovation and the CAMH Research Hospital Fund, was used to perform this research. We thank the NVIDIA Corporation for the Titan Xp GPU that was used for this research. We acknowledge the assistance of the American Association of Suicidology in making the University of Maryland Reddit Suicidality Dataset available. This study was supported by the CAMH Foundation, a National Science and Engineering Research Council of Canada (NSERC) Discovery Grant to LF.

## Conflict of Interest

LF owns shares in Alphabet Inc. which is the parent company of Google, the developer of the freely available Universal Sentence Encoder which was compared to other methods.